\documentclass[letterpaper]{article} 
\usepackage{aaai25}  
\usepackage{times}  
\usepackage{helvet}  
\usepackage{courier}  
\usepackage[hyphens]{url}  
\usepackage{graphicx} 
\usepackage{natbib}  
\usepackage{caption} 
\usepackage{algorithm}
\usepackage{algorithmic}
\usepackage{xcolor}
\usepackage{amsthm}
\usepackage{amssymb}
\usepackage{amsmath}
\usepackage{thmtools}
\usepackage{booktabs} 
\usepackage{subcaption}

\definecolor{new_pink}{HTML}{d59683}
\definecolor{new_blue}{HTML}{1663A9}

\newcommand{\err}[1]{\scriptsize $\pm$ #1}
\usepackage{newfloat}
\usepackage{listings}
\DeclareCaptionStyle{ruled}{labelfont=normalfont,labelsep=colon,strut=off} 
\floatstyle{ruled}
\newfloat{listing}{tb}{lst}{}
\floatname{listing}{Listing}
\title{Differentiable Information Enhanced Model-Based Reinforcement Learning}
\author{
    Xiaoyuan Zhang\textsuperscript{\rm 1,2,3},
    Xinyan Cai\textsuperscript{\rm 4},
    Bo Liu\textsuperscript{\rm 1},
    Weidong Huang\textsuperscript{\rm 3},\\
    Song-Chun Zhu\textsuperscript{\rm 3,1,2},
    Siyuan Qi\textsuperscript{\rm 3},
    Yaodong Yang\textsuperscript{\rm 1,2}\thanks{Corresponding author}
}
\affiliations{
    \textsuperscript{\rm 1} Institute for Artificial Intelligence, Peking University \\
    \textsuperscript{\rm 2} State Key Laboratory of General Artificial Intelligence, Peking University, Beijing, China \\ 
    \textsuperscript{\rm 3} State Key Laboratory of General Artificial Intelligence, BIGAI, Beijing, China \\
    \textsuperscript{\rm 4} Institute of automation, Chinese Academy of Sciences
}

\usepackage{bibentry}

\begin{document}

\maketitle
\begin{abstract}
Differentiable environments have heralded new possibilities for learning control policies by offering rich differentiable information that facilitates gradient-based methods. In comparison to prevailing model-free reinforcement learning approaches, model-based reinforcement learning (MBRL) methods exhibit the potential to effectively harness the power of differentiable information for recovering the underlying physical dynamics. However, this presents two primary challenges: effectively utilizing differentiable information to 1) construct models with more accurate dynamic prediction and 2) enhance the stability of policy training. In this paper, we propose a Differentiable Information Enhanced MBRL method, MB-MIX, to address both challenges. Firstly, we adopt a Sobolev model training approach that penalizes incorrect model gradient outputs, enhancing prediction accuracy and yielding more precise models that faithfully capture system dynamics. Secondly, we introduce mixing lengths of truncated learning windows to reduce the variance in policy gradient estimation, resulting in improved stability during policy learning. To validate the effectiveness of our approach in differentiable environments, we provide theoretical analysis and empirical results. Notably, our approach outperforms previous model-based and model-free methods, in multiple challenging tasks involving controllable rigid robots such as humanoid robots' motion control and deformable object manipulation.
\end{abstract}

\section{Introduction}
    Robot control aims to develop effective control inputs to guide robots in accomplishing assigned tasks. Differentiable environments present a promising opportunity, enabling precise computation of first-order gradients of task rewards with respect to control inputs~\citep{xu2021accelerated}. This gradient information facilitates gradient-based policy optimization methods~\citep{antonova2023rethinking,freeman2021brax}. Model-based reinforcement learning (MBRL) methods show greater potential than model-free approaches in leveraging environmental differentiable information to construct accurate models, thereby  improving performance~\citep{moerland2023model}.
    
    However, the design of MBRL methods in differentiable environments presents new challenges. Firstly, ensuring the stability of policy training in a differentiable environment is crucial. Gradient-based methods have been developed, with the most representative one being SHAC~\citep{xu2021accelerated} (which we will also refer to as a differentiable-based method in the following text). These algorithms make better use of gradient information and often achieve better experimental performance compared to non-differentiable-based approaches(such as PPO~\citep{schulman2017proximal}, SAC~\citep{haarnoja2018soft}, etc.) in differentiable environments. However, due to the presence of collisions(discontinuities in gradients) and the issue of vanishing or exploding gradients during long trajectory back-propagation, differentiable-based methods often suffer from instability of training and are highly sensitive to trajectory lengths~\citep{suh2022differentiable}. This can be particularly detrimental to robot control. For various types of robot control, especially in environments such as humanoid robots, the stability of policy training is crucial for performance~\citep{andrychowicz2020matters,abeyruwan2023sim2real}.Second, in terms of model learning, a critical challenge arises from the accumulation of model errors ~\citep{plaat2023high}, which manifest in both trajectory prediction and first-order gradient prediction~\citep{li2022gradient}. 
    
    In this paper, we present a novel MBRL approach, \textbf{MB-MIX}, to address these challenges. Our method leverages path derivatives~\citep{clavera2020model} for policy optimization. To enhance the stability of policy training, we introduce trajectory length mixing (\textbf{MIX}) in subsection~\ref{sec:model based with mix}. This technique combines both long and short trajectories, reducing variance in policy gradient estimation. We conduct a theoretical analysis of MIX approach in subsection~\ref{sec:analysis}, demonstrating its effectiveness in achieving low-variance policy gradient estimation. To address the problem of model errors, we propose \textbf{Sobolev model training method}~\citep{czarnecki2017sobolev,parag2022value} for learning dynamics models that make effective use of differentiable information.  This approach trains the model by matching its predictions of environmental dynamics and their first-order gradients. We also analyze the internal connection between these two methods in~\ref{sec:Consistent coherence}. By combining trajectory length mixing and Sobolev training, we mitigate the impact of cumulative prediction errors of long trajectory models on policy training.
    
    Finally, we validated the effectiveness of our algorithm in multiple scenarios. We first 
    demonstrated the effectiveness of the MIX approach in a designed simple tabular case experiment. We then conducted experiments on two benchmarks, DiffRL~\citep{xu2021accelerated} and Brax~\citep{freeman2021brax}, which contain classic robot control problems. In DiffRL, our algorithm achieved performance surpassing all state-of-the-art methods. Humanoid robots present a challenging problem in robotics, with bipedal interaction playing a crucial role. These bipedal robots are well-suited for collaboration and work with humans in everyday life and work environments. We innovatively trained a real bipedal humanoid robot, \textbf{Bruce}~\citep{liu2022design}, using differentiable simulation (DiffRL), successfully validating the effectiveness of our algorithm and opening up significant possibilities for future real-robot deployments. In Brax, we proved that utilizing a Sobolev training method for the dynamics model provided greater benefits compared to traditional models. Moreover, in DaXBench\citep{chen2022benchmarking}, we demonstrated the effectiveness of our method in differentiable deformable object environments with large state and action spaces.
    
    In summary, our work makes three main contributions to MBRL in differentiable environments by effectively utilizing differentiable information. \textbf{First}, we propose an MBRL method with trajectory length mixing to reduce variance in policy gradient estimation, thereby achieving more stable training performance. \textbf{Second}, we employ Sobolev training to reduce model errors. This training approach is consistent and coherent with gradient-based policy training methods. \textbf{Third}, our approach achieves better performance compared to state-of-the-art model-free and model-based reinforcement learning methods in multiple robot-control tasks.

\section{Related Work}
\label{sec:citations}
\textbf{Model-Based Reinforcement Learning} methods learn dynamic models to guide policy optimization, reducing sample complexity while maintaining performance. Learned dynamics models fall into two categories: First is enhancing model-free methods with the learned model, enhancing policy optimization through path derivatives~\citep{amos2021model,d2020learn}, where policy gradients are computed via model back-propagation. For example, SVG~\citep{heess2015learning} introduced a framework for learning continuous control policies using model-based back-propagation. MAAC~\citep{clavera2020model} incorporated learned terminal Q-functions to estimate long-term rewards. DDPPO~\citep{li2022gradient} highlights the importance of accurate gradient predictions by the model in influencing policy training. Model-enhanced data methods include MBPO~\citep{janner2019trust}, which trains the SAC algorithm using generated and real trajectories. Similar ideas have been extended to offline model-based RL settings~\citep{yu2020mopo,lee2021representation}. Impressive advancements have also been made in learning dynamic changes in latent variable spaces~\citep{hafner2019learning,hafner2019dream,hafner2020mastering}, and the latest development is DreamerV3~\citep{hafner2023mastering}. Furthermore, the application of transformers as world models~\citep{transformerworldmodel,robine2023transformer100k,ma2024transformer} has demonstrated robust performance in real humanoid robots~\citep{transformerhumanoid}. The second way is to use the model for planning. LOOP~\citep{sikchi2022learning}, TD-MPC~\citep{Hansen2022tdmpc,hansen2024tdmpc2} incorporate terminal value estimates for long-term reward estimates. Sobolev training is a method that improves neural network training by using both function values and derivatives as supervision signals~\citep{czarnecki2017sobolev}. This can enhance prediction accuracy and generalization~\citep{parag2022value} . In this paper, we suggest using the Sobolev training method along with differentiable information from the environment to improve the accuracy of the dynamic model.\\
\textbf{Differentiable Simulators}
are physics engines that can compute gradients of physical quantities concerning simulation parameters or inputs~\cite{newbury2024review}, categorized into rigid-body and soft-body environments~\citep{hu2019taichi,huang2021plasticinelab}. Soft-body simulations, like fluid simulations~\citep{xian2023fluidlab} and deformable objects~\citep{li2023dexdeform,chen2023daxbench}, face challenges like local optima, while rigid-body simulations~\citep{freeman2021brax,xu2021accelerated} confront discontinuous gradients due to collisions~\cite{zhong2022differentiable}. Despite improvements in environment design~\citep{werling2021fast,geilinger2020add}, challenges like collision-induced losses persist~\citep{suh2022differentiable}. In this paper, we have conducted extensive experiments on both rigid robots and soft deformable differentiable environments. The proposed trajectory length mixing method alleviates the gradient impact caused by collisions to some extent.\\
\textbf{Policy Gradient Estimation}
is a technique for optimizing stochastic policies by calculating gradient estimates of expected returns with respect to policy parameters using environment samples~\citep{grondman2012survey,schulman2015trust,schulman2017proximal,haarnoja2018soft}. Differentiable simulators support gradient-based policy optimization by providing gradient information~\citep{du2021diffpd,mora2021pods,freeman2021brax,xu2021accelerated}. Our approach use a hybrid of trajectory lengths and effectively reduce the variance of the policy gradient estimation, improving the stability of policy training.

\begin{figure*}
    \centering
    \includegraphics[width=\textwidth]{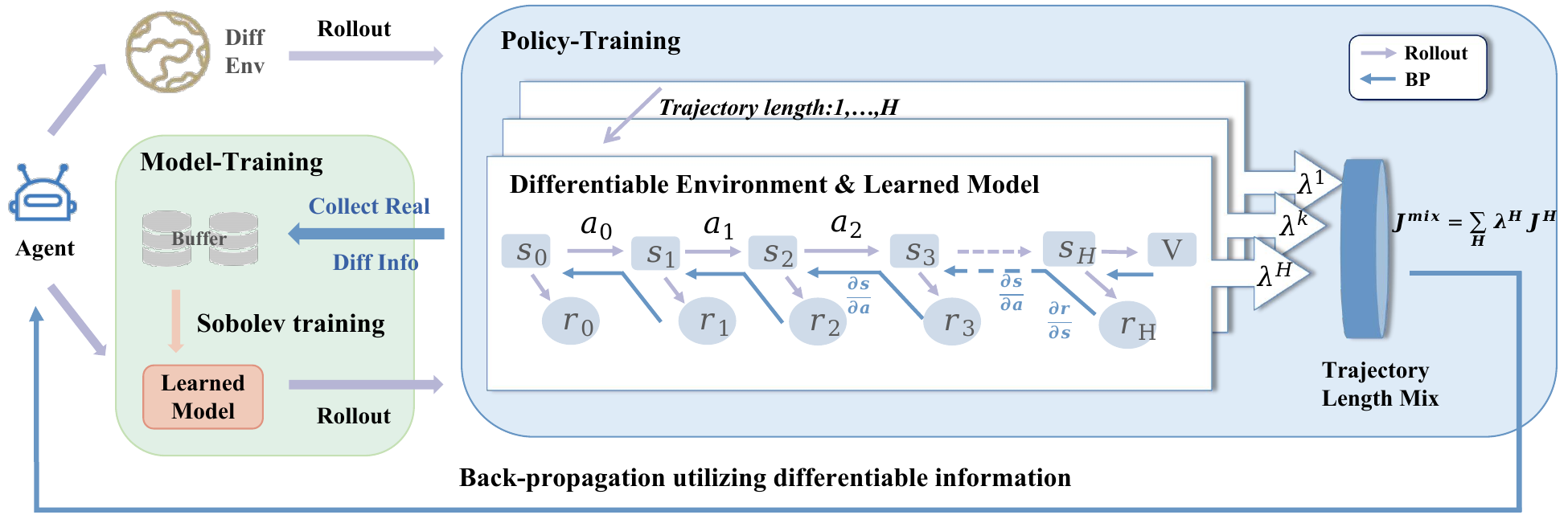}
    \caption{Algorithm diagram. We propose a differentiable information enhanced model-based reinforcement learning approach, \textbf{MB-MIX}, which uses \textcolor{new_pink}{Sobolev model training} method to learn a dynamics model that leverages gradient information(Diff Info) from the differentiable environment. We perform rollouts in the differentiable environment as well as in the learned model, and employ \textcolor{new_blue}{Trajectory Length Mix} to weight and sum the optimization functions. Policy updates are then performed through Back-propagation. It is worth noting that in our method, the \textbf{Model-Training} with Diff Info and the gradient-based \textbf{Policy-Training} method are consistent.}
    \label{fig: Algorithm diagram}
\end{figure*}

\section{Method}
\label{sec:method}
We propose a differentiable information enhanced Model-Based Reinforcement Learning method, MB-MIX, outlined in Figure \ref{fig: Algorithm diagram} and Algorithm~\ref{alg:main}. In subsection \ref{sec:model based with mix}, we present a model-based method that mixes trajectories of different lengths, effectively reducing variance in policy gradient estimation, and mitigating the issue of accumulated model errors in long trajectory predictions in MBRL. In subsection~\ref{sec:Sobolev model training}, we employ the Sobolev training method, which utilizes state transition function gradients in differentiable environments to optimize dynamics model training, and analyze its consistency with MIX methods. In subsection~\ref{sec:analysis}, we provide theoretical analysis demonstrating the MIX method's effectiveness in reducing policy gradient estimation variance.
\subsection{Preliminaries}
We consider a discrete-time infinite-horizon Markov decision process (MDP) defined by a tuple $\left(\mathcal{S}, \mathcal{A}, p_0, f, r, \gamma\right)$, where $\mathcal{S} \subseteq \mathbb{R}^{d_{\boldsymbol{s}}}$ is the space of states, $\mathcal{A} \subseteq \mathbb{R}^{d_{\boldsymbol{a}}}$ the space of actions, $p_0\left(\boldsymbol{s}_0\right)$ the distribution over initial states $\boldsymbol{s}_0$, $\boldsymbol{s}_{t+1} \sim f\left(\boldsymbol{s}_t, \boldsymbol{a}_t\right)$ the transition function, $r\left(\boldsymbol{s}_t, \boldsymbol{a}_t\right)$ the reward function, and $\gamma \in(0,1]$ a discount factor. A trajectory $\tau=\left(\boldsymbol{s}_0, \boldsymbol{a}_0, \boldsymbol{s}_1, \boldsymbol{a}_1, \ldots, \boldsymbol{s}_H, \boldsymbol{a}_H \right) \in \mathcal{T}$, is a sequence of states and actions of length $H$ sampled from the dynamics defined by the MDP and the policy, where the horizon $H$ may be infinite. $\mathcal{T} \subseteq \mathbb{R}^{(d_{\boldsymbol{s}} + d_{\boldsymbol{a}}) \times (H+1)}$ is the space of trajectories and $\pi(\boldsymbol{a}_t | \boldsymbol{s}_t;\theta)$ is the policy parameterized by $\boldsymbol{\theta} \in \mathbb{R}^{d_{\boldsymbol{\theta}}}$. The objective of an RL algorithm is to train a policy that maximizes the expected sum of discounted rewards: $J(\theta) = \mathbb{E}_{\tau \sim p(\tau; \theta)} \left[\sum_{t=0}^{H} \gamma^t r(\boldsymbol{s}_t, \boldsymbol{a}_t)\right]$. A value function $V(\boldsymbol{s})$ is often learned to approximate $J(\theta)$. The input-output relationship of dynamic model M is defined as $\hat{s}_t = M(s_{t-1}, a_{t-1})$.

\subsection{Model-Based Approach with Mixed Trajectory Length}
\label{sec:model based with mix}
Commonly used policy update approaches using path derivatives often collect trajectories of a specific length for reward aggregation, which is used to propagate updates to the policy. In particular, the general form of the objective function can be written as:
\begin{small}
\begin{equation}
\label{eq:1}
J_{\pi}^H(\theta) = E\left[\sum\nolimits_{t=0}^{H-1}\gamma^t r(s_t,a_t) + \gamma^H V(s_H)\right].
\end{equation}
\end{small}
In practice, varying trajectory lengths $H$ significantly impact policy gradient estimation. 
Moreover, model-based methods can face increased cumulative error with longer trajectory predictions. To address these challenges, we propose trajectory length mixing. This technique blends trajectories of different lengths to reduce policy gradient estimation variance and improve model training stability. Specifically, it computes the optimization function by weighted averaging expected reward aggregation from trajectories of varying lengths:
\begin{small}
\begin{equation}
\label{eq:2}
J_{\pi}^{\text{mix}}(\theta)=(1-\lambda)\sum\nolimits_{H=1}^{\infty}\lambda^{H-1}J_{\pi}^H(\theta).
\end{equation}
\end{small}
In practice, the trajectory length is upper-bounded by the upper limit of a specific task or simply by choosing a higher length. Additionally, the interval of mixing trajectories of different lengths is adjustable. By substituting Equation~\ref{eq:1} into Equation~\ref{eq:2}, we have: 
\begin{small}
\begin{equation}
\label{eq:3}
J^{\text{mix}}_{\pi}(\theta)=E\left[\sum\nolimits_{t=0}^{\infty}\left((\gamma\lambda)^t (r(s_{t},a_{t})+(1-\lambda)\gamma V(s_{t+1})\right)\right]
\end{equation}
\end{small}
The transition of the state function is predicted by a learned dynamics model. Here, we adopt the idea of branch roll out~\citep{janner2019trust}, starting from states in the real environment and rolling out in the model to control the impact of model errors. For policy gradient estimation, the model accuracy has a greater impact on long trajectories, whereas the accuracy of the value function plays a more substantial role for shorter trajectories.
Furthermore, Equation~\ref{eq:3} represents a general form of the optimization function for gradient-based policy training using path derivatives. When $\lambda=1$, $J^{\text{mix}}_{\pi}(\theta)=E\left[\sum_{t=0}^{\infty}(\gamma)^t (r(s_{t},a_{t})\right]$     equivalent to $SVG(\infty)$\citep{heess2015learning}. When $\lambda = 0$, $J^{\text{mix}}_{\pi}(\theta)=E\left[(r(s_{t},a_{t})+\gamma V(s_{t+1}))\right]$ equivalent to $SVG(1)$.

\subsection{Sobolev Model Training Method}
\label{sec:Sobolev model training}
To leverage the rich gradient information in differentiable environments, we propose using the Sobolev training method to learn the environment dynamics model. Sobolev training incorporates the gradient information into the loss function, thereby encouraging the model to learn more about the gradients of the input values during the training process. The loss function is:
\begin{small}
\begin{equation}
\label{eq:4}
\begin{split}
J_{M}(\varphi)= & E\bigg[\|\widehat{s}_{t+1}-s_{t+1}\|_2+\alpha \bigg[\|\frac{\partial \widehat{s}_{t+1}}{\partial s_t}-\frac{\partial s_{t+1}}{\partial s_t}\|_2 \\
& +\|\frac{\partial  \widehat{s}_{t+1}}{\partial a_t}-\frac{\partial s_{t+1}}{\partial a_t}\|_2\bigg]\bigg].
\end{split}
\end{equation}
\end{small}
where $\varphi $ represents the parameters of the dynamics model $M$, and $\hat{s}$ represents the predicted state.
The dynamics model trained using the Sobolev method maintains consistency with the path gradient-based policy training approach described in subsection~\ref{sec:model based with mix}. This can be observed from the chain rule formula for gradient back-propagation in Equation~\ref{eq:5}, a detailed expansion of $J^{\text{mix}}$ is included in Appendix:
\begin{small}
\begin{equation}
\label{eq:5}
\begin{split}
\frac{\partial J_{\pi}}{\partial \theta}= & \mathbb{E}\left[\sum_{i=0}^\infty\frac{ \partial J_{\pi}}{\partial a_i}\frac{\partial a_i}{\partial \theta}\right]=\mathbb{E}\sum_{i=0}^\infty\bigg(\gamma^i\frac{\partial r_i}{\partial a_i} \\
& +\sum_{k=i+1}^\infty\gamma^k \frac{\partial r_k}{ \partial s_k}\frac{ \partial s_k}{\partial s_{i+1}}\frac{\partial s_{i+1}}{\partial a_i}\bigg)\frac{\partial \pi_{\theta}(s_i)}{\partial \theta}.
\end{split}
\end{equation}
\end{small}
\textbf{Consistency between model training and policy training: }
\label{sec:Consistent coherence}
\textbf{MIX} (Using trajectories of different lengths) is helpful for stability, but in the model-based setting, longer trajectories suffer from accumulated errors. In model-based policy training methods utilizing path derivatives, the dynamics model must accurately predict the zeroth-order values of $(s,a)$ and provide precise predictions for the gradients of the transition function. Previous model-based approaches prioritized accurate zeroth-order values over first-order gradients. Dynamics models trained using Sobolev training methods emphasize the accuracy of both zeroth-order and first-order values. Therefore, to utilize longer trajectories in the MBRL setting, \textbf{Sobolev Model training method} is needed to reduce the error in gradient predictions. We list the pseudo code in Algorithm~\ref{alg:main}.

\begin{algorithm}[t]
\caption{MB-MIX}  
\label{alg:main}
\begin{algorithmic}[1]
   \STATE {\bfseries Input:} policy $\pi_{\boldsymbol{\theta}}$, Model $M_{\varphi}$, Value function $V_{\psi}$, Environment buffer $D_{env}$
   \FOR{$i=0,....,K-1$}
   \STATE{Sample trajectories {$s_t,a_t,r_t,s_{t+1},\frac{\partial s_{t+1}}{\partial s_t},\frac{\partial s_{t+1}}{\partial a_t}$} from real environment with $\pi_{\boldsymbol{\theta}}$ add to $D_{env}$.}
   \STATE{Train $M_{\varphi}$ using data from $D_{env}$ to minimize $J_M(\varphi)$}
   \FOR{$t=0,1,2,...,T$}
   \STATE sample state, action, reward trajectories through the interaction of policy $\pi_{\boldsymbol{\theta}}$ and $M_{\varphi}$ 
   \STATE{compute $J^{\text{mix}}_{\pi}(\boldsymbol{\theta})$ for all time steps.}
   \STATE{$\boldsymbol{\theta} = \boldsymbol{\theta} - \alpha \nabla_{\boldsymbol{\theta}} J^{\text{mix}}_{\pi}(\boldsymbol{\theta})$.}
   \STATE{$\hat{V}(s_t)=(1-\lambda)(\sum\limits_{k=1}^{h-t-1}\lambda^{k-1}G_t^k)+\lambda^{h-t-1}G_t^{h-t}$, where $G_t^K=\sum\limits_{l=0}^{k-1}\gamma^lr(s_{t+l})+\gamma^kV(s_{t+k})$}
   \STATE{Train $V_{\psi}$  to minimize $L=E_{s\in \tau_i}[||V_{\psi}(s)-\hat{V}(s)||].$}
   \ENDFOR
   \ENDFOR
   \RETURN{Optimal policy $\pi_{\boldsymbol{\theta}}$}
\end{algorithmic}
\end{algorithm}


\begin{figure*}[t]
\centering
\includegraphics[width=1.0\linewidth]{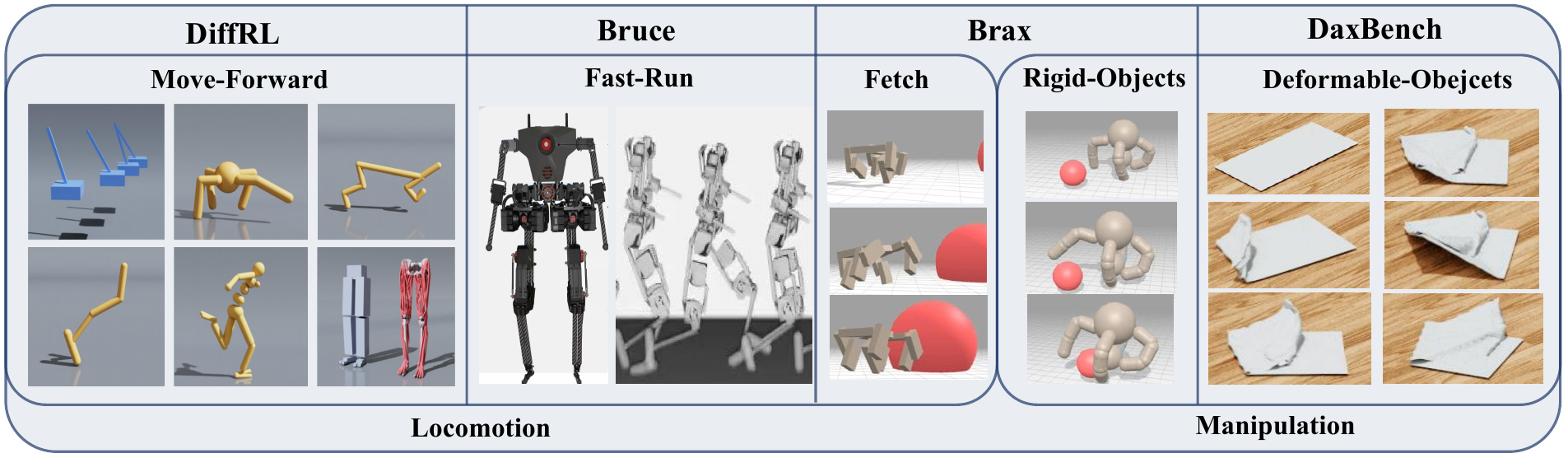}
\caption{\textbf{Task Gallery. }We assessed the algorithm's effectiveness in four differentiable environments, encompassing various control tasks.
(a) \textbf{DiffRL:} The agent controls a range of robots, including those with muscles.
(b) \textbf{Bruce, Humanoid Robot:} Designed for tasks like Fast-Run, was introduced into the DiffRL environment to extend its real-world applications.
(c) \textbf{Brax:} Involving advanced tasks such as Fetch and Grasp
(d) \textbf{DaxBench:} Entailing a series of tasks related to deformable objects manipulation.} 
\label{fig:experiment}
\end{figure*}
\subsection{Theoretical Analysis}
\label{sec:analysis}
In the subsequent analysis, we delve into the Variance for the MIX estimator and the SHAC method(a representative differentiable based method) using a differentiable environment scenario. Based on some common assumptions detailed in Appendix Section A (We consider these assumptions to be common constraints for proving convergence and bounds as shown in~\citep{fallah2021convergence,clavera2020model}), we deduce an upper bound for the Variance of these estimators and provide the necessary conditions for this bound to hold. A key finding from our analysis is that the MIX estimator of the policy gradient has an existing Variance bound, and this bound is lower than that of the SHAC method for any trajectory length under balance factor $\lambda$ and discount factor $\gamma$ less than 1. This result suggests a performance enhancement for the MIX estimator.

\begin{restatable}{theorem}{mixmcq_modelfree}\label{mse_bound_modelfree}
Suppose Assumptions $1, 2$, $3$, and $4$ hold (please see appendix). Then in a differentiable environment, the Variance of the MIX policy gradient estimate ($A^{\textup{MIX}}$) will be equal to or less than the Variance of the SHAC policy gradient estimate ($A^{\textup{SHAC}}$):
\begin{equation}
\label{eq:MSE}
\operatorname{Variance}\left(A^{\textup{MIX}}\right) \leq \operatorname{Variance}\left(A^{\textup{SHAC}}\right)
\end{equation}
Here, $A^{\text{MIX}}$ and $A^{\text{SHAC}}$ respectively denote the MIX and SHAC policy gradient estimates,   the balance factor \(\lambda\) , the discount factor \(\gamma\) are positive real numbers less than 1.
\end{restatable}
Using all of the assumptions, we bound the variance term for MIX and SHAC. The complete proof with all steps and mathematical details can be found in Appendix Section A. 


\section{Experimental Results}
\label{sec:result}
Our experiment aims to investigate the following questions: (1) Can trajectory length mixing improve gradient-based policy optimization methods? Especially in terms of training stability. (2) How does the proposed MB-MIX approach perform compared to state-of-the-art reinforcement learning methods? (3) Does the use of a dynamics model trained with Sobolev improve the performance of model-based reinforcement learning methods? We conducted experiments in the environment depicted in Figure~\ref{fig:experiment} to address the aforementioned questions:\\
1. We developed a straightforward tabular case environment with discrete state and action spaces. Additionally, experimental evaluations were performed on DaXBench, a sophisticated deformable object manipulation environment~\citep{chen2022benchmarking}. Through these experiments, we provide empirical evidence for the efficacy of trajectory length mixing. \\
2. We conducted extensive experiments in the differentiable and parallelizable robot control environment, DiffRL~\citep{xu2021accelerated}. Our approach was compared against state-of-the-art gradient-based, model-free, and model-based methods, outperforming them across various tasks. Ablation experiments were also performed to analyze the individual effects of each component. Excitingly, we introduced Bruce~\citep{liu2022design}, a real bipedal humanoid robot, into the DiffRL environment. Bruce's real-world nature introduced additional constraints and collisions, demanding robustness and posing new challenges in policy training. Remarkably, our algorithm achieved performance surpassing all SOTA baselines, greatly enhancing the potential for deploying it on real robots. \\
3. We validated the effectiveness of our proposed Sobolev model training approach in the Brax environment, as mentioned in subsection~\ref{sec:Sobolev model training}. Models trained with the Sobolev training approach exhibited better performance in assisting gradient-based policy training.

\subsection{Tabular Case: Effectiveness of MIX}
\begin{figure}
  \centering
  \includegraphics[width=0.47\textwidth]{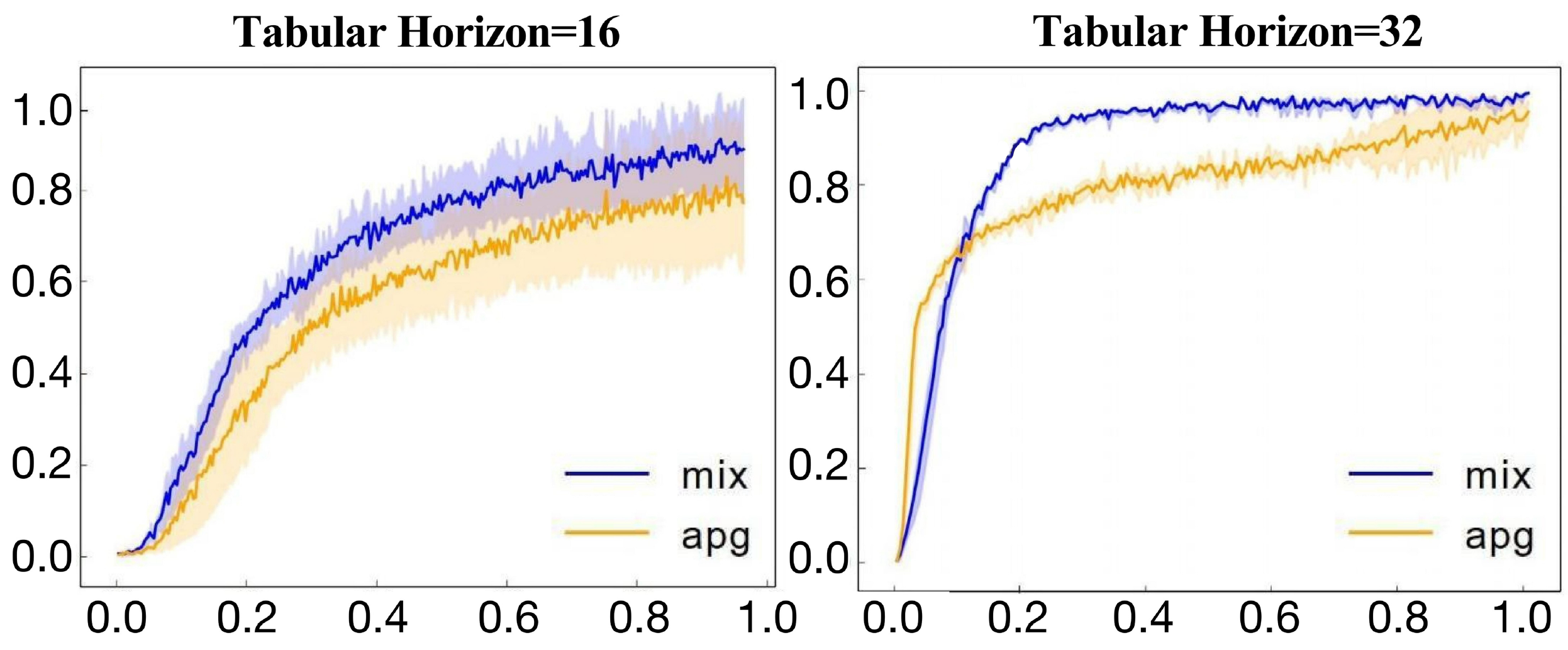}
  \caption{Experiment results in Tabular case. We show the effectiveness of mixing trajectory length in a designed simple tabular case environment. The legend "reward/step" on the y-axis denotes the average reward. The right end of the horizontal axis represents 1e6 environment steps.}
\label{fig:tabular}
\end{figure}
\textbf{Training Setting: } We consider a tabular random Markov Decision Process setting as described in~\citep{liu2022theoretical}. The state and action spaces have dimensions 20 and 5, respectively, yielding a reward matrix $R \in \mathbb{R}^{20\times5}$. The transition probability matrix is generated from independent Dirichlet distributions. The initial policy is a matrix $\theta^{0}\in \mathbb{R}^{20\times5}$, and the final policy $\pi_{\theta}$ is obtained via softmax activation: $\pi_{\theta}(a| s)=\exp (\theta(s, a)) / \sum_{b} \exp (\theta(s, b))$. The tabular MDP provides a controlled setting to evaluate the effectiveness of trajectory length mixing. The training objective was to optimize the policy through improved gradient-based methods. The experimental settings involved varying trajectory lengths, using a $\lambda = 0.98$ discount rate for reward-to-go calculations, and adjusting the mix-interval accordingly.\\
\textbf{Main Result: }As shown in Figure~\ref{fig:tabular}, our MIX approach outperforms baseline methods in terms of convergence speed and final policy performance. 
The baseline APG~\citep{wiedemann2023training} uses complete trajectory lengths for policy training. Without additional noise factors, this experiment indicates that trajectory length mixing can enhance the performance of gradient-based policy optimization.

\begin{table*}[t]
\centering
\resizebox{1.0\textwidth}{!}
{
\begin{tabular}{c c c c c c c c c}
    \hline
        & \multicolumn{3}{c}{Model-Free Method} & \multicolumn{4}{c}{Model-Based Method}\\ \cmidrule(lr){2-4} \cmidrule(lr){5-8}
        & SHAC & PPO & SAC & DreamerV3 & LOOP & MAAC & \textbf{MB-MIX(ours)} \\ \hline
        
        Ant & 5174\err{349} & 1213\err{217} & 1278\err{39} & 1989\err{374} & 2995\err{249} & 5289\err{651} & \textbf{6363}\err{27} \\
        
        Cheetah & \textbf{8366}\err{259} & 5552\err{661} & 6392\err{215} & 8034\err{142} & 7029\err{468} & 7832\err{239} & \textbf{8483}\err{80} \\
        
        Hopper & 3638\err{222} & 382\err{71} & 384\err{34} & 3010\err{722} & 542\err{297} & 2056\err{317} & \textbf{4048}\err{127}\\
        
        Cartpole & -622\err{25} &  -1794\err{108} & -1607\err{76} & -666\err{59} & -801\err{76} & -629\err{32} & \textbf{-604}\err{9} \\ 
        
        Humanoid & 3321\err{968} & 279\err{104} & 889\err{41} & 294\err{103} & 369\err{38} & 1297\err{387} & \textbf{4955}\err{197}\\
        
        SNU-Humanoid & 3722\err{395} & 88\err{2} & 49\err{7} & 35\err{5} & 26\err{10} & 2774\err{444} & \textbf{3907}\err{94} \\

    \hline
    \end{tabular}
}
\caption{\textbf{Experiment results in DiffRL.} In a differentiable robot control environment, our method outperforms state-of-the-art model-free and model-based methods, demonstrating the overall effectiveness.}
\label{table: whole}
\end{table*}

\subsection{DiffRL: Comparison with SOTA methods}
\begin{figure}
\includegraphics[trim={3pt 0pt 0pt 0pt},clip,width=0.47\textwidth]{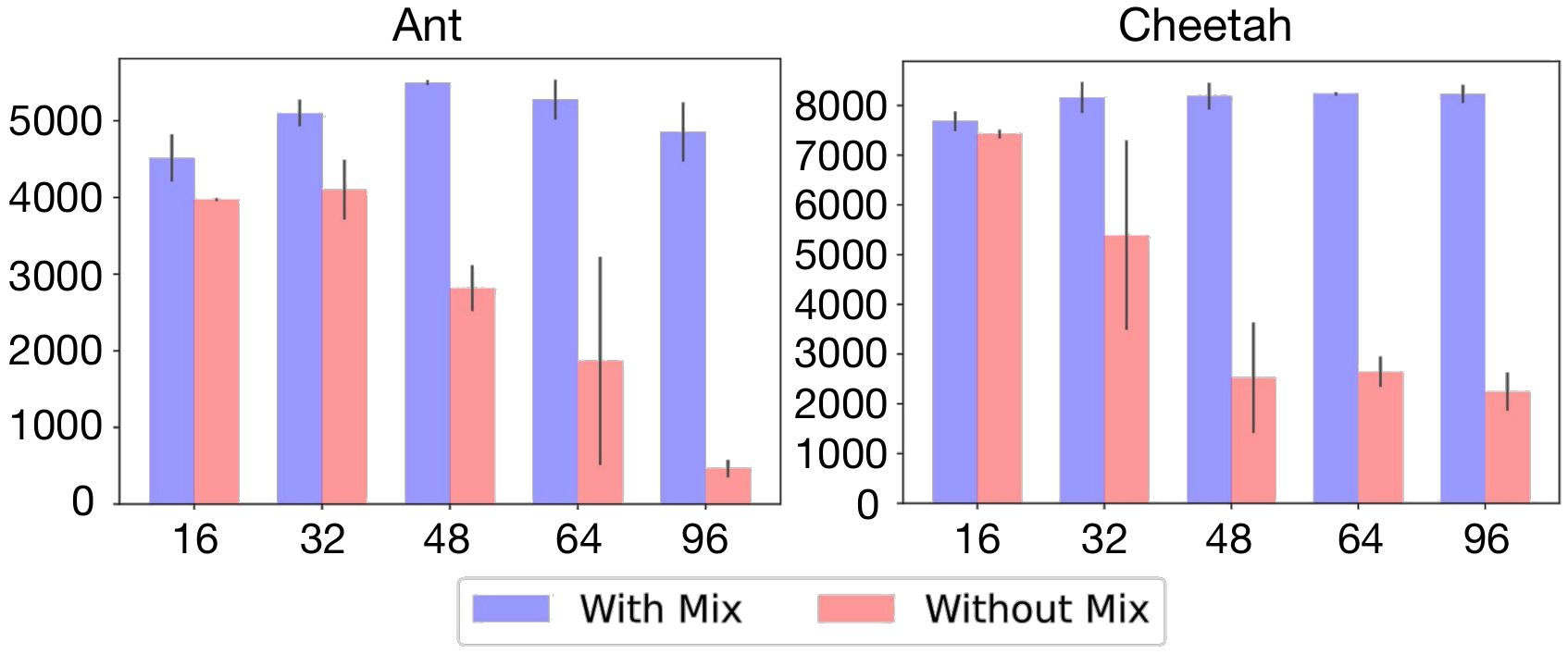}
  \caption{Impact of different trajectory lengths on training.  The vertical axis of the figure represents reward, while the horizontal axis represents the maximum length of trajectories. Our Mix method enhances policy training stability.}
  \label{fig:different length}
\end{figure}
\textbf{Training Setting: }We conduct experiments in the differentiable parallel environment DiffRL~\citep{xu2021accelerated}, which includes tasks such as Ant, Humanoid, and SNU-Humanoid. The objective was reward maximization. Lower parallel environments (4 and 8) were used to highlight sample efficiency of model-based methods. In the experiment, our MB-MIX algorithm was trained on all six tasks, with a $ \lambda = 0.98$ and mix-interval set to 1 or 2 depending on the task. \\
\textbf{Compared Baseline: } The sample efficiency and numerical performance of the proposed MB-MIX algorithm were compared with state-of-the-art model-free and model-based RL methods, including path-derivative-based (e.g., SHAC~\citep{xu2021accelerated}) and non-path-derivative-based (e.g., PPO~\citep{schulman2017proximal}, SAC~\citep{haarnoja2018soft}) model-free methods, as well as model-based methods like DreamerV3~\citep{hafner2023mastering}, LOOP~\citep{sikchi2022learning}, and MAAC~\citep{clavera2020model}.\\
\textbf{Main Result: } Table~\ref{table: whole} explicitly demonstrates that our MB-MIX method outperforms all baselines on all six tasks. Particularly, we achieve remarkably impressive performance on the Humanoid and ant tasks. It is worth noting that our algorithm, compared to the state-of-the-art gradient-based method SHAC in differentiable environments, not only achieves better performance but also demonstrates higher stability across all tasks. \\
\textbf{Ablation Experiment:} We conducted experiments on the stability of the trajectory mixing methods. We compared the impact of different trajectory lengths on training in the Ant and Cheetah environments, as illustrated in Figure~\ref{fig:different length}. The x-axis of the graph represents the maximum trajectory length used for each update (the length of interaction between the agent and the environment). The results showed that, under different trajectory lengths, Our \textbf{Mix} Method achieved more stable and better policy training performance. 
\subsection{Bruce: Humanoid Robotic Control}
\begin{figure*}[t]
\centering
\includegraphics[width=1.0\linewidth]{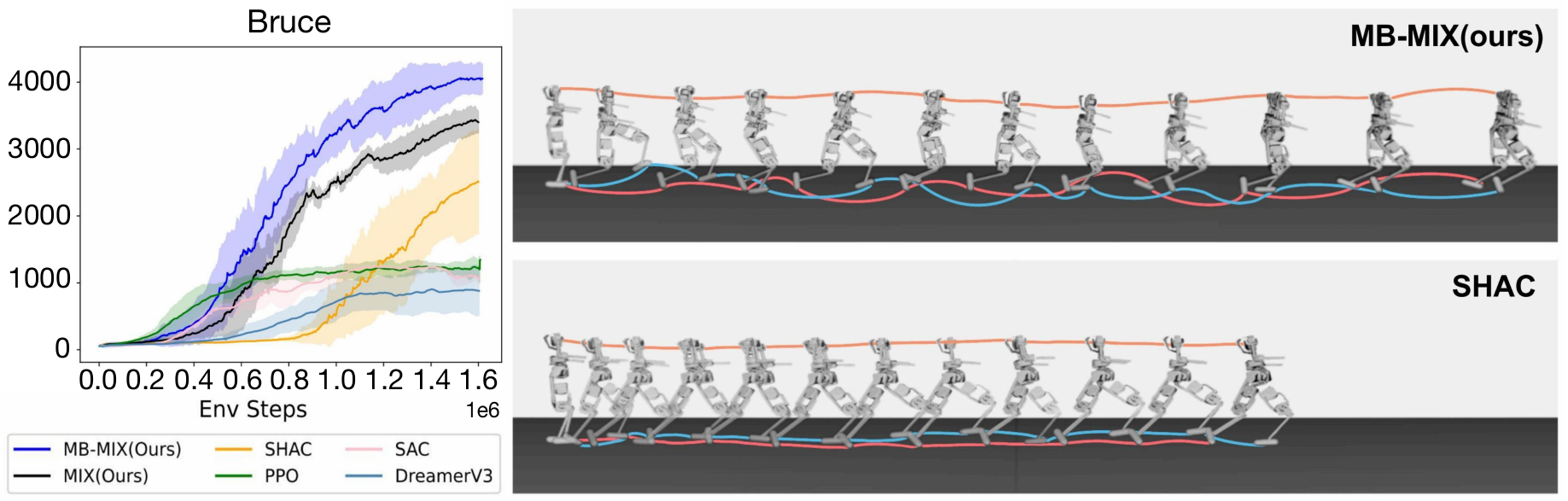}
\caption{Experiments on Bruce, humanoid robot. The top-left figure demonstrates that our proposed MB-MIX method surpasses all model-free and model-based algorithms, the vertical axis represents the rewards. In the top-right figure, we visualize the performance in the “Fast Run” task. Starting from the same position and after the same amount of time, our MB-MIX method enables the trained robot to move further. Judging from the alternation of the red and blue lines, which track the footsteps, our MB-MIX algorithm achieves better alternating leg movements.} 
\label{fig:bruce experiment}
\end{figure*}

\begin{figure}
  \includegraphics[trim={0pt 0pt 0pt 0pt},clip,width=0.44\textwidth]{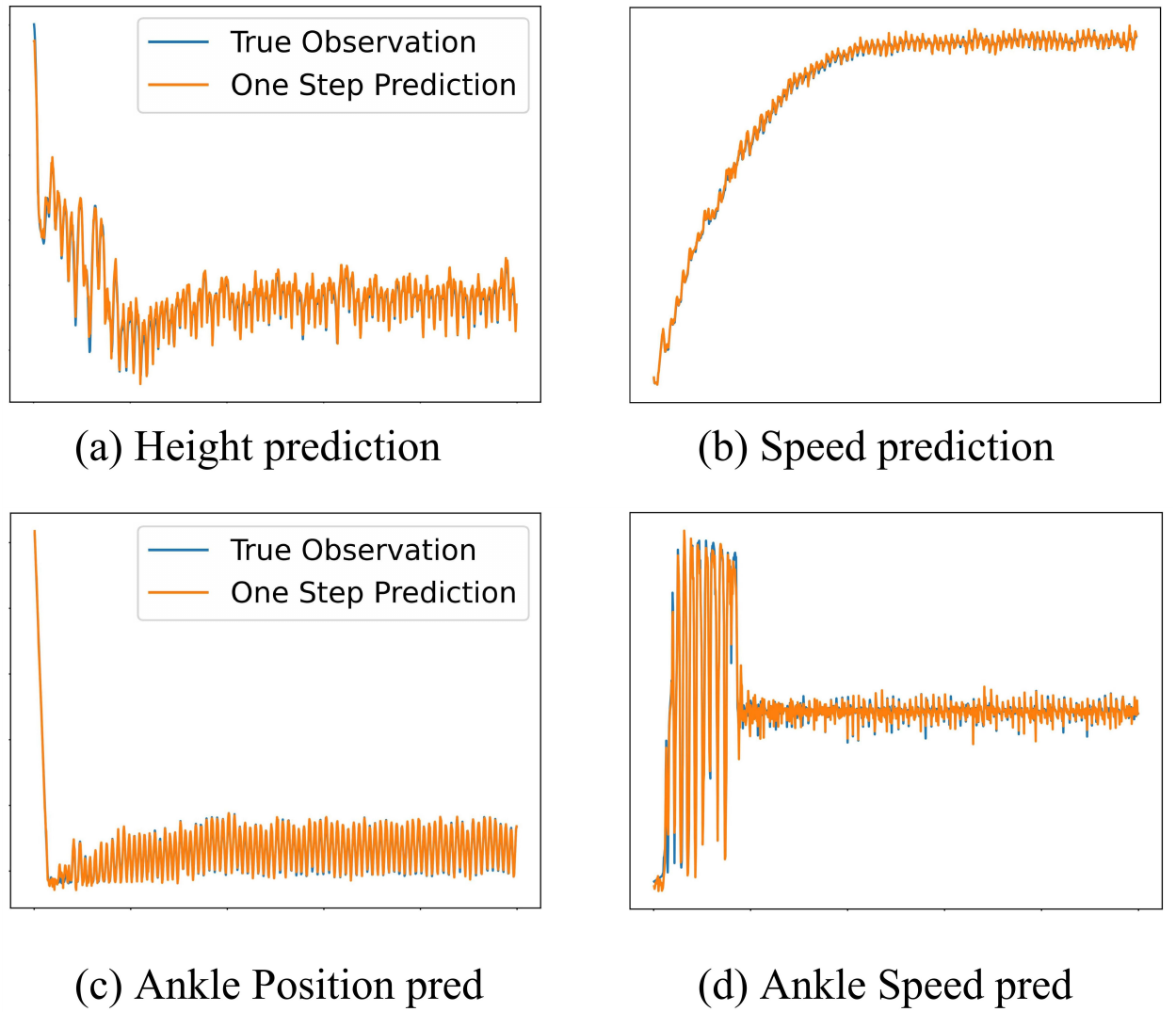}
  \caption{Learned model predictions on Bruce: The state predicted by the model and the actual state. The horizontal axis represents the environmental steps. The vertical axis represents the corresponding predicted values: Height, Speed, Ankle Position, Ankle Speed.}
  \label{fig:learned model prediction}
  \vspace{-10pt}
\end{figure}

\textbf{Training Setting: }Bruce is a miniature bipedal robot with five degrees of freedom (DoF) per leg, including a spherical hip joint, a knee joint, and an ankle joint. To reduce leg inertia, Bruce employs cable-driven differential pulley systems and link mechanisms for the hip and ankle joints, respectively, enabling a human-like range of motion~\citep{liu2022design}. "Fast Run" task evaluates robot's forward speed.\\
\textbf{Main Result: }The algorithm's performance on the "Fast Run" task is shown in Figure~\ref{fig:bruce experiment}. Our MB-MIX and MIX methods outperform other approaches, exhibiting higher reward and greater stability, consistent with theoretical analysis. Notably, \textbf{MB-MIX} demonstrates significant improvement over \textbf{MIX}, indicating the learned Dynamic Model effectively aids policy learning, serving as an ablation experiment for the dynamic model. The right side of Figure~\ref{fig:bruce experiment} visualizes the policy performance of \textbf{MB-MIX} versus the SHAC algorithm. Starting from the same initial position and duration, MB-MIX enables the robot to travel farther and perform better in the "Fast Run" task. The MB-MIX-trained robot exhibits alternating fast forward motion, which is more "human-like" and efficient compared to SHAC. Figure~\ref{fig:learned model prediction} demonstrates the learned model's effectiveness in state prediction, with the dynamic model accurately aligning with the ground truth. This provides strong support for reducing wear and tear during real-world robot interactions. Experiment indicates our algorithm's tremendous potential for bipedal robot applications.

\subsection{Brax: Model Training Performance}
\label{sec:brax}
\textbf{Training Setting: }Brax is a environment with interactive objects. Tasks in Brax include Fetch and Grasp, which have continuous state and action spaces, providing us with more challenging test beds.  We used a dynamics model trained with Sobolev training, which allows us to compare the effectiveness of this method against traditional ones.\\

\begin{figure}
  \includegraphics[trim={3pt 0pt 0pt 0pt},clip,width=0.44\textwidth]{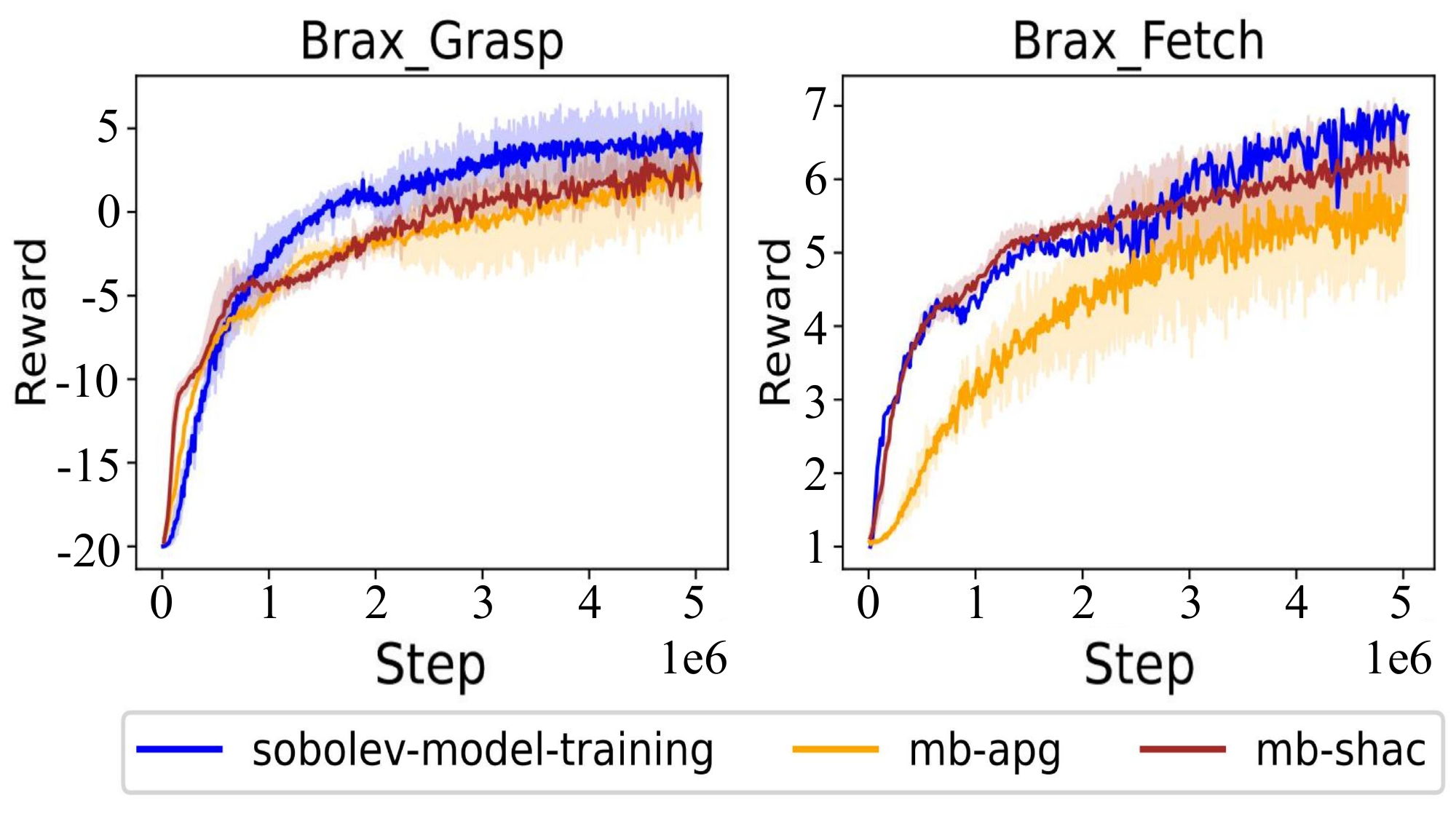}
  \caption{Experiments on the Brax benchmark. In a more complex robot control environment with Fetch and Grasp tasks, sobolev-model-training  method enhances the overall performance.}
  \label{fig:Brax}
  \vspace{-20pt}
\end{figure}

\textbf{Main Result: }The results illustrated in Figure~\ref{fig:Brax} demonstrate the superior performance of our approach, which combines Sobolev model training method for the dynamics model, surpassing the benchmark methods. Notably, the performance gap becomes more pronounced with increasing task complexity, underscoring the effectiveness and scalability of our sobolev model training approach. These findings affirm that the integration of a Sobolev-trained dynamics model can indeed enhance the performance of MBRL methods.

\subsection{DaXBench: Efficiency on Deformable Objects with Large State-Action Spaces}
\label{sec:daxbench}
\begin{table}
\resizebox{0.47\textwidth}{!}{
\begin{tabular}{lcccc}    
\toprule    Task & APG & SHAC & \textbf{MIX} \\
\midrule    
Fold-Cloth-1 &  0.36\err{0.06} & 0.34\err{0.07} &  \textbf{0.52\err{0.02}} \\
Fold-Cloth-3 &  \textbf{0.19\err{0.09}} & \textbf{0.22\err{0.22}} & \textbf{0.23\err{0.08}}  \\
Unfold-Cloth-1 &  0.42\err{0.02}  & 0.50 \err{0.03} & \textbf{0.73\err{0.01}} \\
Unfold-Cloth-3 &  0.39\err{0.02}  & \textbf{0.48 \err{0.03}} & \textbf{0.51\err{0.03}}\\
\bottomrule   
\end{tabular} 
}
\caption{\textbf{DaXBench}. In the deformable objects environment, our proposed  MIX approach achieved better performance compared to gradient-based algorithms}   
\vspace{-15pt}
\label{table: DaxBench}
\end{table}
\textbf{Training Setting: }DaXBench~\citep{chen2022benchmarking} is a high-performance differentiable simulation platform, ideal for deformable object manipulation (DOM) research. Tasks in DaXBench cover deformable objects and manipulation tasks, providing our method with more challenging test beds. We conducted experiments to evaluate the stability of the trajectory mixing methods within this environment.\\ \textbf{Main Result: }The results in Table~\ref{table: DaxBench} demonstrate MIX approach leads to high performance and low variance across all tasks, outperforming the benchmark methods. Indicating the effectiveness and scalability of our approach in diverse and complex tasks. These findings confirm that the MIX approach can indeed enhance the performance of DOM methods, providing a stable and efficient solution.

\section{Conclusion}
\label{sec:conclusion}
We propose a novel MBRL framework for differentiable environments. The method introduces a trajectory length mixing technique to mitigate the impact of varying trajectory lengths on policy gradient estimation, thereby enhancing the stability of policy training. Additionally, we innovatively leverage the Sobolev method to learn accurate dynamics models that align with the gradient-based policy training. Theoretical analysis validates the advantages of the trajectory length mixing technique. Experimental evaluation on rigid robot control and deformable object manipulation tasks demonstrates superior performance over prior model-based and model-free approaches.
\section{Limitations and Future Work}
The theoretical analysis is limited. We focus on the MIX method's impact on policy gradient estimation, yet have not extended the analysis to the model-based setting. Future research could explore the model-based configuration.

\bibliography{main}

\end{document}